\documentclass[sigconf]{acmart}
\AtBeginDocument{%
  \providecommand\BibTeX{{%
    \normalfont B\kern-0.5em{\scshape i\kern-0.25em b}\kern-0.8em\TeX}}}
\setcopyright{acmcopyright}
\copyrightyear{2020}
\acmYear{2020}
\acmDOI{10.1145/1122445.1122456}

\acmConference[ACMMM 2021]{Proceedings of the 29th ACM International Conference on Multimedia}{20-24 October 2021}{Chengdu, China}
\acmBooktitle{Proceedings of the 29th ACM International Conference on Multimedia (MM '21),
20-24 October 2021 Chengdu, China}
\acmPrice{15.00}
\acmISBN{978-1-4503-XXXX-X/18/06}

\usepackage{times}
\usepackage{epsfig}
\usepackage{graphicx}
\usepackage{algorithm}
\usepackage{algorithmicx}
\usepackage{algpseudocode}
\usepackage{amsmath}
\usepackage{multirow}
\usepackage{setspace}
\usepackage{diagbox}
\begin{document}
\title{An Effective and Robust Detector for Logo Detection}

\author{Xiaojun Jia}
\authornote{Both authors contributed equally to this research.}
\email{jiaxiaojun@iie.ac.cn}
\affiliation{%
  \institution{Institute of Information Engineering, Chinese Academy of Sciences}
  \state{Beijing}
  \country{China}
}
\author{Huanqian Yan}
\authornotemark[1]
\email{yanhq@buaa.edu.cn}
\affiliation{%
 \institution{Beijing Key Laboratory of Digital Media, School of Computer Science and Engineering, Beihang University}
  \state{Beijing}
  \country{China}
 }
\author{Yonglin Wu}
\authornotemark[1]
\email{shadowuyl@foxmail.com}
\affiliation{%
  \institution{Institute of Information Engineering, Chinese Academy of Sciences}
  \state{Beijing}
  \country{China}
}
\author{Xingxing Wei}
\authornote{Corresponding author.}
\email{xxwei@buaa.edu.cn}
\affiliation{%
 \institution{Institute of Artificial Intelligence, Hangzhou Innovation Institute, Beihang University}
  \state{Beijing}
  \country{China}
 }
\author{Xiaochun Cao}
\authornotemark[2]
\email{caoxiaochun@iie.ac.cn}
\affiliation{%
  \institution{Institute of Information Engineering, Chinese Academy of Sciences}
  \state{Beijing}
  \country{China}
}
\author{Yong Zhang}
\email{zhangyong201303@gmail.com}
\affiliation{%
 \institution{Institute of Automation, Chinese Academy of Sciences}
  \state{Beijing}
  \country{China}
 }

\renewcommand{\shortauthors}{Jia, Yan, and Wu, et al.}

\begin{abstract}
In recent years, intellectual property (IP), which represents literary, inventions, artistic works, etc, gradually attract more and more people's attention. Particularly, with the rise of e-commerce, the IP not only represents the product design and brands, but also represents the images/videos displayed on e-commerce platforms. Unfortunately, some attackers adopt some adversarial methods to fool the  well-trained logo detection model for infringement. To overcome this problem, a novel logo detector based on the mechanism of \textit{looking and thinking twice} is proposed in this paper for robust logo detection. The proposed detector is different from other mainstream detectors, which can effectively detect small objects, long-tail objects, and is robust to adversarial images. In detail, we extend \textit{detectoRS} algorithm to a cascade schema with an equalization loss function, multi-scale transformations, and adversarial data augmentation. A series of experimental results have shown that the proposed method can effectively improve the robustness of the detection model. Moreover, we have applied the proposed methods to competition ACM MM2021 \textit{Robust Logo Detection} that is organized by Alibaba on the Tianchi platform and won top 2 in 36489 teams. Code is available at \textit{https://github.com/jiaxiaojunQAQ/Robust-Logo-Detection}.
\end{abstract}


\maketitle

\section{Introduction}
On e-commerce platforms, commodity logo is one of the intellectual property rights of commodities \cite{machado2015brand, bossel2019facing}. It represents the creations and ideas of the business owners. However, some illegal merchants adopt adversarial attack methods to fool the well-trained logo detection model for infringement \cite{trappey2021intelligent}. Adversarial examples \cite{szegedy2013intriguing,zhao2020object}, which are generated via adding indistinguishable perturbations to benign images, can fool the well-trained logo detection model. Hence, we need a robust logo detection model to protect the intellectual property (IP). Logo detection belongs to object detection which is used to recognize and locate objects. A series of object detection methods \cite{ren2015faster,cai2018cascade, qiao2021detectors, hu2021a2} have been proposed in recent years. These object detectors have achieved excellent performance on some conventional benchmarks such as PASCAL VOC \cite{everingham2010pascal} and MS COCO \cite{lin2014microsoft}. But these methods are not suitable for natural images with a long-tailed Zipfian distribution in a realistic scenario. Moreover, these models are easily attacked by adversarial examples. As for the logo detection in the realistic scenario, the logo detection model not only needs to detect and recognize logos but also needs to defend against adversarial examples. In detail, the logo detection task has following challenges. 1). small object detection, 2). long tail category detection and 3). robust detection with adversarial images.
 
 \begin{figure*}[t]
\begin{center}
   \includegraphics[width=0.95\linewidth]{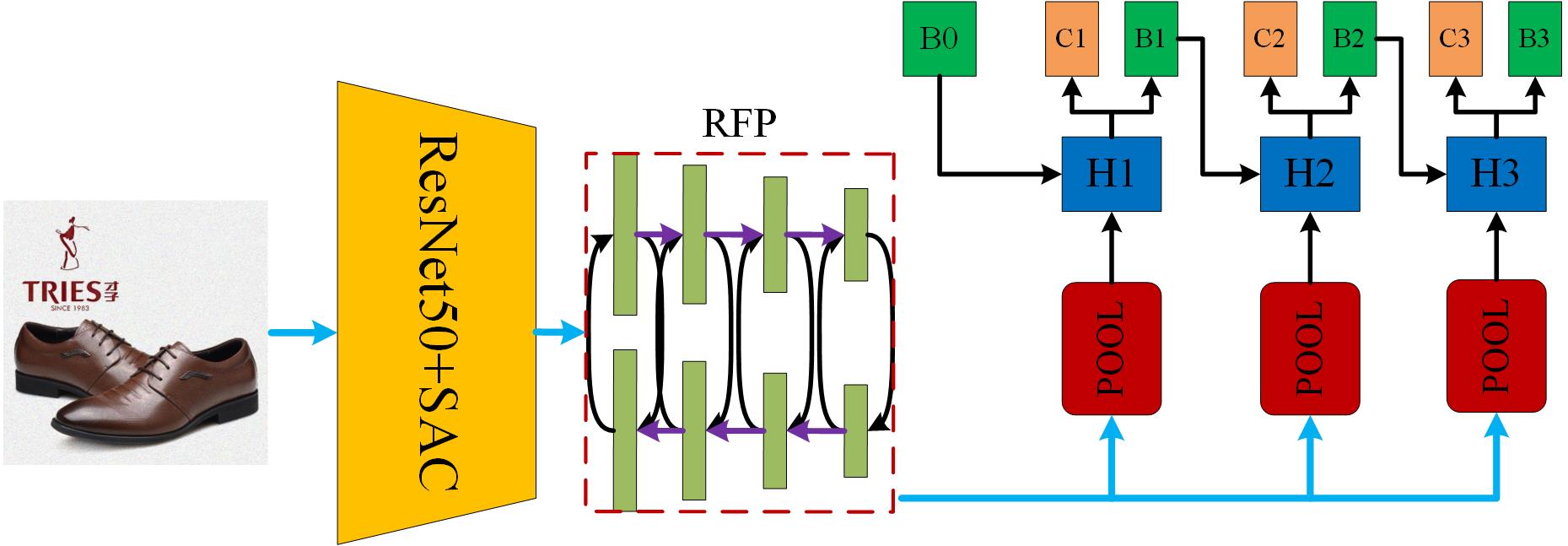}
\end{center}
   \caption{The network structure of the Cascade detectoRS. SAC (switchable Atrous Convolution) represents switchable atrous convolution, which can adaptively select the receptive field. RFP (Recursive Feature Pyramid) adopts a cyclic structure to repeatedly utilize and refine the extracted features. ``H'' represents the detection head, ``B'' represents the regression box, ``C'' represents the category prediction result, and ``B0'' represents the output result of the RPN network.
   }
\label{fig:acm_mm_01}
\end{figure*}
 \par Small object detection is one of the difficulties in the field of object detection. For a realistic scenario logo detection, the average logo pixel is less than 1\% of the whole image pixel. For a long tail category distribution in the realistic scenario, the natural images belong to a long-tailed Zipfian distribution, whose annotations of tail classes are insufficient for training a detector. For robust detection with adversarial images in a realistic scenario, it is very important. Because some illegal merchants adopt adversarial attack methods to fool the well-trained logo detection model for infringement. Hence, a robust detection model that can deal with adversarial images well is crucial. To our best knowledge, there are few studies on the robustness of the object detection models against adversarial examples. But the research about the robustness of the logo detection against adversarial examples is of great importance in a realistic scenario.

\par To overcome the above problems, a novel logo detector based on the mechanism of looking and thinking twice is proposed in this paper. More concretely, we extend \textit{detectoRS} algorithm to a cascade schema with an equalization loss function, multi-scale transforms, and adversarial data augmentation. We train our logo detection model on the Logo detection database Open Brand \cite{jin2020open}, which consists of 584,920 natural images with 1,303,563 annotations. We propose to use the multi-scale training and testing strategy to improve the model's ability to detect small objects. As for the long tail category detection, we adopt the  equalization loss v2 (EQL-v2) \cite{tan2021equalization} to protect the learning of tail categories. And as for the detection robustness against adversarial examples, we propose to use adversarial data augmentation to improve the robustness of the  detector. Moreover, we find that the multi-scale training and testing strategy not only improves the ability to detect small objects but also improves the robustness against adversarial examples. The detection framework is shown in Fig. \ref{fig:acm_mm_01}. Our main contributions can be summarized as follows:
\begin{itemize}
    \item We propose a novel logo detector based on the mechanism of \textit{looking and thinking twice} to improve the robustness against adversarial images.
    \item We find that the multi-scale training and testing strategy not only improves the ability to detect small objects but also improves the robustness against adversarial examples.
    \item A  series of experimental results show that the proposed method can improve the robustness of the detector in the realistic scenario.
\end{itemize}
The rest of this paper is organized as follows. The proposed algorithms are described in Section 2. The experimental results and analysis are presented in Section 3. Finally, we summarize the work in Section 4.
\section{Methodology}
In this section. We first introduce the logo detection model proposed in this paper. And then we will focus on the three modules designed for solving the problems mentioned earlier.
\subsection{Logo detection model}
An effective detection model plays an important role in defending against adversarial images and detection performance. We extend the original detectoRS model to cascade schema in our experiments. The reason we select detectoRS and cascade is that these two methods have same mechanism \emph{looking and thinking twice}, and this mechanism is similar to the human vision system. Therefore, we think this combination may be robust to various adversarial perturbations. The network structure is shown in Fig.~\ref{fig:acm_mm_01}. In detail, we adopt the Resnet 50 \cite{he2016deep} as the backbone network. The Recursive Feature Pyramid (RFP) is used as the neck layer, which can reuse and refine the extracted features. It can be defined as:
\begin{equation}
    f_{i} = F_{i}(f_{i+1},x_{i}),\qquad x_{i} = B_{i}(x_{i-1},R_{i}(f_i))
\end{equation}
where $B_i$ denotes the $i_{th}$ stage of the bottom-up backbone, $F_i$ denotes the $i_{th}$ top-down FPN \cite{ren2017faster} operation, $R_i$ denotes the feature transformations before connecting them back to the bottom-up backbone, $f_i$ is the feature map, and $x$ is the input image. Additionally, We also use the Switchable Atrous Convolution (SAC) to adaptively select the receptive field and an attention-based feature fusion mechanism which is the same as the global context module of SeNet \cite{hu2018squeeze}, to enhance the network's representation ability in the proposed detector.

\subsection{Multi-scale training and testing strategy}
We adopt the multi-scale training and testing strategy, which is a simple but effective method. We choose this strategy on the one hand to solve the problem of small objects, on the other hand to improve the robustness of the model. For small object detection, one of the most intuitive solutions is to expand the input size of the network. To not destroy the receptive field of the detection model, multi-scale training and testing strategy is used in our experiments. In this way, while taking into account the problem of small object detection, we also improve the detection performance on the normal size object detection. In the experiment, we set the maximum side length of the training and testing scale to 1333, and the short side scale ranges from 800 to 1100. Moreover, we find that the multi-scale training and testing strategy also can improve the robustness against adversarial examples. It is because that multi-scale image transformation can 
break up the particular structure of the adversarial perturbations. Some studies \cite{song2017pixeldefend, jia2019comdefend} have proved that breaking up the adversarial perturbation structure is an effective method to defend against adversarial examples.
\subsection{Equalization loss}
The cross-entropy loss has an inhibitory effect on non-ground-truth classes. When calculating the gradient, it will produce a negative gradient with an inhibitory effect in other categories. And the long tail category is affected by the negative gradient of other categories, and the accumulation will be greater than the positive gradient generated by itself. It will be hard for the detection model to learn the features of the long tail category. A simple but effective method is to directly ignore the loss that has a negative effect on the long tail categories when calculating loss. In this way, the detection model can learn the better features of long tail categories. It can be simply defined as:
\begin{equation}
    L_{EQL}=-\sum_{j=1}^{C}W_{j}*log(p_j),
\end{equation}
where $C$ represents the number of categories, $W_{j}$ represents the weight of the category $j$, and $p_j$ represents the predicted probability of the current proposal category by the network. 

But this loss function ignores the influence of the background candidate area. To overcome this problem, we use equalization loss \cite{tan2021equalization} as the classification loss function in the detection model. Equalization loss is based on gradient guidance. It weights the positive and negative gradients according to the cumulative ratio of the positive gradient and the negative gradient. The positive weights $q_j^{(t)}$ and negative weights $r_j^{(t)}$ can be defined as:
\begin{equation}
   \begin{cases}
q_j^{(t)} = 1+4*(1-f(g_j^{(t)}))\\
r_j^{(t)} = f(g_j^{(t)}),
\end{cases}
\end{equation}
where $t$ represents the number of iterations and 
$f(x)=1 /(1+e^{-12 *(x-0.8)})$.
\par After the positive and negative gradient weights are obtained, the current gradient values can be updated separately.
They can defined as:
\begin{equation}
\begin{cases}
\bigtriangledown_{z_j}^{pos^{\prime}}(L^{(t)})=q_j^{(t)}\bigtriangledown_{z_j}^{pos}(L^{(t)})\\
\bigtriangledown_{z_j}^{neg^{\prime}}(L^{(t)})=q_j^{(t)}\bigtriangledown_{z_j}^{neg}(L^{(t)}),
\end{cases}
\end{equation}
where ${z_j}$ represents the output of the classifier, and $L$ represents the loss function. Moreover, the cumulative ratio of positive and negative gradients for the $t+1$ time can be defined as $g_j^{t+1} = \sum_{t=0}^{T}|\bigtriangledown_{z_j}^{pos^{\prime}}(L^{(t)})|/\sum_{t=0}^{T}|\bigtriangledown_{z_j}^{neg^{\prime}}(L^{(t)})$
\subsection{ Adversarial data augmentation}
In addition to using the input multi-scale transformation strategy, we also simulate the generation of adversarial attack noise and enhance the robustness of the detection model by augmenting the adversarial data. The illegal merchants usually generate adversarial examples to fool the well-trained logo detection model for infringement under unrestricted attack in the realistic scenario. Considering this, we first generate adversarial examples by adding Gaussian noise, adding rain, adding fog, and image blurring, etc. And then we inject these adversarial examples into the training data set and train the detection model. In this way, it can improve the robustness against adversarial images. 
\begin{figure*}[t]
  \centering
  \includegraphics[width=0.85\linewidth]{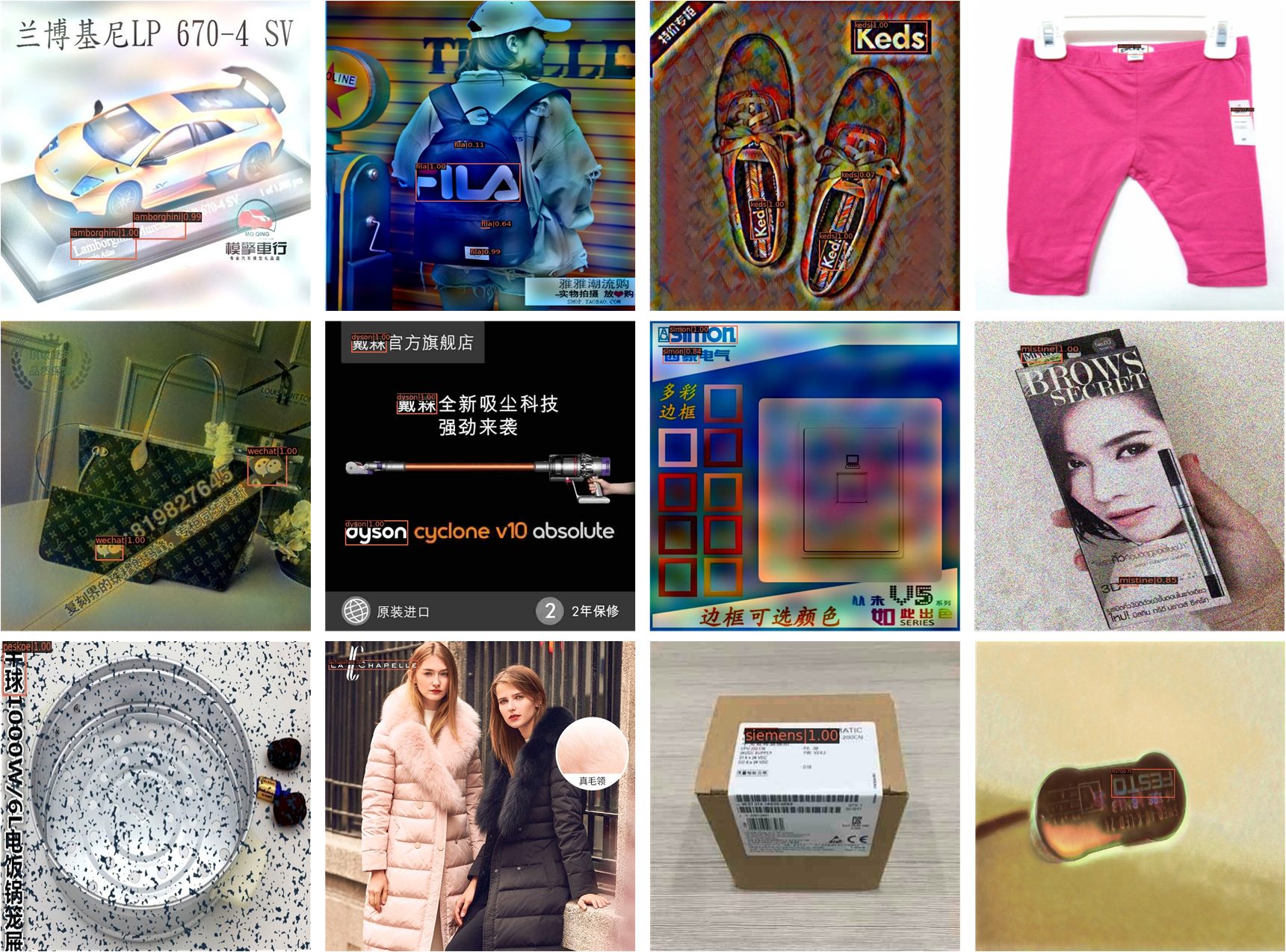}
  \caption{Some detection results with adversarial logo images. These images show obvious signs of manipulation from various unknown adversarial operations that would disable the detector. It is no surprise that our detector is somewhat defensible against those apparent interference.}
  \label{imgresults}
\end{figure*}

\section{Experiment}
We used the proposed detector in ACM MM2021 Security AI Challenger Phase 7: \textit{ Robust Logo Detection} competition, and got evaluation index $MAP$ 0.650807, which is the second highest result submitted by all the participating teams. Some detection images are shown in Fig. \ref{imgresults}. These images show obvious signs of manipulation from various unknown adversarial operations that would disable the detector. It is no surprise that our detector is somewhat defensible against those apparent perturbations. The result of the competition can show the rationality and effectiveness of our method. In the following content, some ablation experiments have been done for analyzing and recording the detection performance changes of our algorithm.

In the competition, the indicator $MAP$ is used to evaluate the detection performance. The index $MAP$ is the average of $AP$ metrics for all detection categories. The $AP$ computes the average value of $Precision$ over the interval from $Recall=0$ to $Recall=1$, i.e., the area under the $PRC$, so the higher $AP$ value is, the better the detection performance and vice versa. A detection is considered to be true positive if the area overlap ratio $r$
between the predicted bounding box and the ground truth bounding box exceeds a presetting threshold value. In order to evaluate the detection performance more strictly and accurately, more thresholds are selected in the final evaluation index, like $\lbrace 0.5:0.05:0.95 \rbrace$.

\begin{figure}[t]
  \centering
  \includegraphics[width=0.95\linewidth]{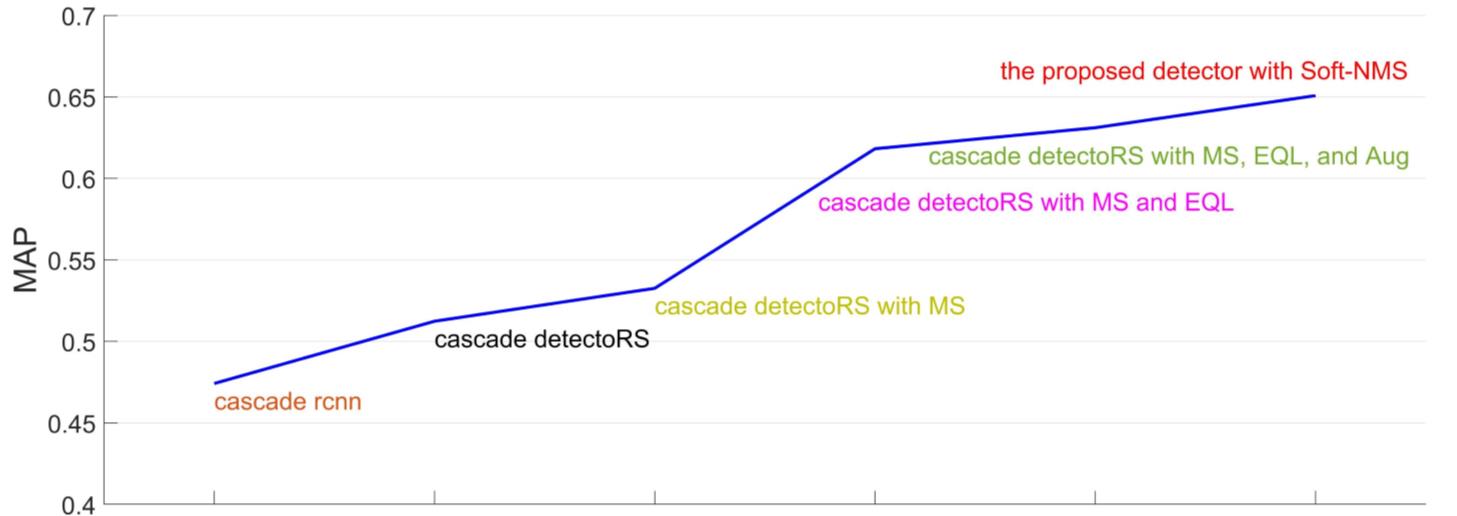}
  \caption{The six detector results with the test dataset. The cascade detectoRS means cascade RCNN with SAC and RFP modules. The cascade detectoRS with MS means detectoRS with multi-scale training and testing. The cascade detectoRS with MS and EQL means cascade detectoRS is improved with multi-scales and equalization loss. The signs "Aug" and "soft-NMS" mean adding data augmentation operations and replacing standard-NMS with Soft-NMS.}
  \label{resultsplot}
\end{figure}

There are 50472 test images, which are polluted with image disturbances. As shown in Fig. \ref{imgresults}, the adversarial perturbations are obvious, these adversarial images are robust and can make common detectors invalid, such as Faster RCNN \cite{ren2017faster}, RetinaNet \cite{lin2020focal}, Libra RCNN \cite{pang2019libra} and so on. To show the effectiveness of the proposed detector, the ablation studies of Multi-scale Training/Testing, Equalization Loss, and Data Augmentation are shown in Fig.\ref{resultsplot}. There are five variant methods: cascade RCNN with backbone 50 \cite{cai2018cascade}, cascade RCNN with sac and RFP (cascade detectoRS), cascade detectoRS with multi scale, cascade detectoRS with multi scales and equalization loss, cascade detectoRS with multi scales, equalization loss, and data augmentation. We train all models for 20 epochs with a learning rate multiplied by 0.01 after 8, 12, and 16 epochs.

It is obvious that all our improvements are effective. The cascade detectoRS is about 8\% ahead than cascade RCNN. Mutil-Scale training and testing can lead to an approximately 3.9\% increase in detection performance. It is amazing, equalization loss brings about 16.1\% performance improvement. The improvement effect of data augmentation is less obvious than other schemes, but it is effective. In addition, replacing standard NMS with Soft-NMS \cite{bodla2017soft} in our experiments also has an obvious promotion. We guess that Soft-NMS can retain more small objects, thus it can improve the recall values. Therefore, we can conclude that the proposed detector is effective, and it can deal with adversarial images well.
\section{Conclusion}
In this paper, a novel logo detector based on the mechanism of \emph{looking and thinking twice} is proposed for robust logo detection against adversarial images in a  realistic scenario. Specifically, we adopt three effective strategies to overcome the difficulties that the logo detection model meets in the realistic scenario. In detail, the multi-scale training and testing strategy is used to improve the model’s ability to detect small objects. Equalization loss is used to protect the learning of tail categories. And the adversarial data augmentation is used to improve the robustness against adversarial examples. Moreover, we find that the multi-scale training and testing strategy also can improve the robustness of the detector. A series of experimental results show that the proposed method can effectively improve the robustness of the detector.
\bibliographystyle{ACM-Reference-Format}
\bibliography{ref}
\end{document}